%% file: main.tex
\pdfoutput=1
\documentclass[letterpaper, 10 pt, conference]{ieeeconf}  % Comment this line out if you need a4paper

\usepackage{blindtext, graphicx}
\usepackage{gensymb}
\usepackage{xcolor}

\usepackage{tikz,lipsum}
\usepackage[labelformat=simple]{subcaption}

\DeclareCaptionLabelSeparator{periodspace}{.\quad}
\usepackage{listings}

\usepackage{algorithm}
\usepackage{algorithmic}

\usepackage{ amssymb }
\usepackage{amsmath}
\usepackage{commath}

\usepackage{framed} 

\usepackage{float}
\usepackage{url}

\usepackage{array}
\newcolumntype{P}[1]{>{\centering\arraybackslash}p{#1}}
\newcolumntype{M}[1]{>{\centering\arraybackslash}m{#1}}

\newcommand*{\prob}{\mathsf{P}}

\IEEEoverridecommandlockouts                              % This command is only needed if 
\title{\LARGE \bf
Two-Stage Transfer Learning for Heterogeneous Robot Detection and 3D Joint Position Estimation in a 2D Camera Image Using CNN
}

\author{Justinas Mi\v{s}eikis$^{1}$, Inka Brija\v{c}ak$^{2}$, Saeed Yahyanejad$^{3}$, Kyrre Glette$^{4}$, Ole Jakob Elle$^{5}$, Jim Torresen$^{6}$% <-this % stops a space
\thanks{$^{1}$ $^{4}$ $^{5}$ $^{6}$Justinas Mi\v{s}eikis, Kyrre Glette, Ole Jakob Elle and Jim Torresen are with the Department of Informatics, University of Oslo, Oslo, Norway}
\thanks{$^{2}$ $^{3}$ Inka Brija\v{c}ak and Saeed Yahyanejad are with the Joanneum Research - Robotics, Klagenfurt am W\"orthersee, Austria}
\thanks{$^{4}$ $^{6}$Kyrre Glette and Jim Torresen also have affiliation with RITMO, University of Oslo}
\thanks{$^{5}$Ole Jakob Elle has his main affiliation with The Intervention Centre, Oslo University Hospital, Oslo, Norway {\tt\small oelle@ous-hf.no}}%
\thanks{$^{1}$ $^{4}$ $^{6}$ {\tt\small \{justinm,kyrrehg,jimtoer\}@ifi.uio.no}}%
\thanks{$^{2}$ {\tt\small Inka.Brijacak@joanneum.at}}%
\thanks{$^{3}$ {\tt\small Saeed.Yahyanejad@joanneum.at}}%
}

\begin{document}

\maketitle
\thispagestyle{empty}
\pagestyle{empty}

%%%%%%%%%%%%%%%%%%%%%%%%%%%%%%%%%%%%%%%%%%%%%%%%%%%%%%%%%%%%%%%%%%%%%%%%%%%%%%%%
\begin{abstract}

Collaborative robots are becoming more common on factory floors as well as regular environments, however, their safety still is not a fully solved issue. Collision detection does not always perform as expected and collision avoidance is still an active research area. Collision avoidance works well for fixed robot-camera setups, however, if they are shifted around, Eye-to-Hand calibration becomes invalid making it difficult to accurately run many of the existing collision avoidance algorithms. We approach the problem by presenting a stand-alone system capable of detecting the robot and estimating its position, including individual joints, by using a simple 2D colour image as an input, where no Eye-to-Hand calibration is needed. As an extension of previous work, a two-stage transfer learning approach is used to re-train a multi-objective convolutional neural network (CNN) to allow it to be used with heterogeneous robot arms. Our method is capable of detecting the robot in real-time and new robot types can be added by having significantly smaller training datasets compared to the requirements of a fully trained network. We present data collection approach, the structure of the multi-objective CNN, the two-stage transfer learning training and test results by using real robots from Universal Robots, Kuka, and Franka Emika. Eventually, we analyse possible application areas of our method together with the possible improvements.

\end{abstract}

\input{main_text}

\section*{ACKNOWLEDGMENT}
This work is partially supported by The Research Council of Norway as a part of the Engineering Predictability with Embodied Cognition (EPEC) project, under grant agreement 240862, and Research Council of Norway through its Centres of Excellence scheme, project number 262762, and by the Austrian Ministry for Transport, Innovation and Technology (BMVIT) within the project framework CollRob (Collaborative Robotics).

% Commented out to make the references nicely fit into one page
%\addtolength{\textheight}{-12cm}   % This command serves to balance the column lengths
                                  % on the last page of the document manually. It shortens
                                  % the textheight of the last page by a suitable amount.
                                  % This command does not take effect until the next page
                                  % so it should come on the page before the last. Make
                                  % sure that you do not shorten the textheight too much.

%%%%%%%%%%%%%%%%%%%%%%%%%%%%%%%%%%%%%%%%%%%%%%%%%%%%%%%%%%%%%%%%%%%%%%%%%%%%%%%%

%%%%%%%%%%%%%%%%%%%%%%%%%%%%%%%%%%%%%%%%%%%%%%%%%%%%%%%%%%%%%%%%%%%%%%%%%%%%%%%%

%%%%%%%%%%%%%%%%%%%%%%%%%%%%%%%%%%%%%%%%%%%%%%%%%%%%%%%%%%%%%%%%%%%%%%%%%%%%%%%%
%\section*{APPENDIX}

%\section*{ACKNOWLEDGMENT}

%%%%%%%%%%%%%%%%%%%%%%%%%%%%%%%%%%%%%%%%%%%%%%%%%%%%%%%%%%%%%%%%%%%%%%%%%%%%%%%%
% \clearpage
\bibliographystyle{IEEEtran}
\bibliography{IEEEexample}

\end{document}

%% file: main_text.tex
% {\colour{red} RED TEXT --- JUSTAS WILL RE-WRITE --- }

% {\colour{blue} BLUE TEXT --- SAEED AND INKA PLEASE RE-WRITE --- IT WOULD BE GREAT TO GET EDITS AND INPUT FOR INTRODUCTION PART ALSO }

\section{INTRODUCTION}

Collaborative robots are gaining popularity as an advanced version of traditional industrial robots. Not only they are capable of reliably performing high-precision complex movements repetitively without any fatigue or rest, but they are also claimed to be safe to operate around humans. Instead of fully separating them from people (e.g., using fences or light curtains), they are capable of sharing the same workspace with humans given the sophisticated collision detection systems. However, these systems do not always work as expected and might exert excess forces before stopping~\cite{ur5collision}. Furthermore, in some situations, like a robot located in a surgery theatre, collisions are not acceptable, and full collision avoidance should be implemented. This coincides with the goals of the Industry 4.0 concept~\cite{lee2015cyber}.

A crucial part for the obstacle avoidance is getting real-time measurements of the workspace and environment around the robot. Such sensing can be done using a variety of sensors: laser scanners, mono and stereo vision, RGB-D cameras, ultrasound sensors and motion capture systems. %Each one has its own pros and cons, and typically a mixture of them are needed to ensure a robust operation.

\begin{figure}[ht]
% \vspace{0.2cm}
\centering
\begin{subfigure}[t]{0.15\textwidth}
    \includegraphics[width=\textwidth]{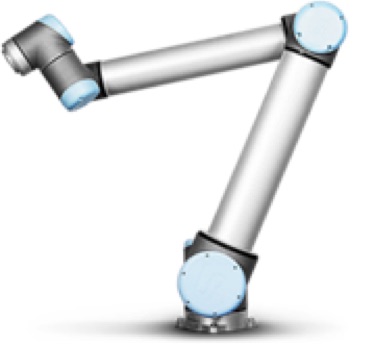}
    \caption{UR3, UR5,  UR10}
    \label{fig:ur}
\end{subfigure}
\hfill
\begin{subfigure}[t]{0.15\textwidth}
    \includegraphics[width=\textwidth]{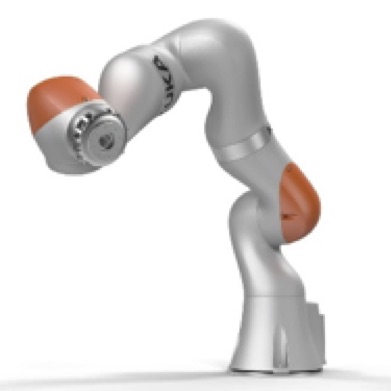}
    \caption{KUKA LBR iiwa}
    \label{fig:kuka}
\end{subfigure}
\hfill
\begin{subfigure}[t]{0.17\textwidth}
    \includegraphics[width=\textwidth]{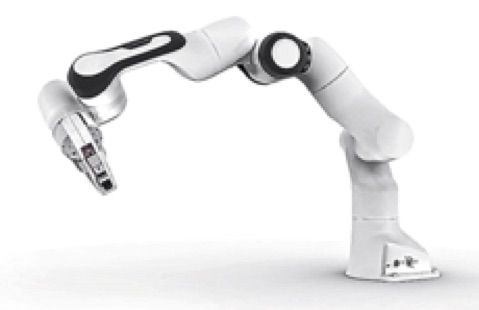}
    \caption{Franka Emika Panda}
    \label{fig:franka_panda}
\end{subfigure}
\caption{Robot manipulators used in our experiments.}
\label{fig:robots_used}
\vspace{-0.4cm}
\end{figure}

Even with advanced sensing systems, the problem still stands in the requirement of having a reliable calibration between the sensors and the robot - so-called Eye-to-Hand calibration~\cite{lenz1989calibrating}. Such a calibration maps the coordinate frames of the robot and vision sensors into a common coordinate frame. As a result, the position of an obstacle detected by one of the sensors can be easily calculated in point of view of the robot, and the necessary action is taken to avoid it. There are reliable and even automatic ways of performing Eye-to-Hand calibration, however, if any of the sensors is unexpectedly moved in relation to the robot, and unaccounted for, the calibration becomes invalid and the system might malfunction~\cite{miseikis2016automatic}. This can be an issue in dynamic environments like a surgery theatre, where there is a lot of human movement, as well as the equipment is constantly shifted around. Similar works and research on dynamic obstacle avoidance for robot arms normally require a fully-calibrated robot-camera system, which can be a challenge in non-static configuration setups~\cite{mainprice2013human}~\cite{miseikis2016multi}.

We have shown that the transfer learning approach can be used to adapt the system trained to recognise and estimate the position of the robot base and joints from one robot model to a new unseen one by having a limited amount of data~\cite{miseikis2018multi}~\cite{miseikis2018transfer}.
We base our work on a previously trained multi-objective CNN on Universal Robots (UR) and extend our work in the following manner. Instead of adapting the network to the new robot type, we adjust the CNN to incorporate new robot types, while still being able to recognise previously trained robots. Eventually, the proposed system is capable of identifying 5 different robots. Furthermore, with the help of motion capture system tracking the camera, we collected a complex training datasets with the camera being moved around in an unconstrained manner, obtaining a variety of viewing angles of the robots in front of complex backgrounds. A more thorough analysis also shows the impact of the accuracy depending on the distance between the camera and the robot.

This paper is organized as follows. First, we provide an overview of related work in Section~\ref{sec:related_work}. We present the system setup and dataset collection in Section~\ref{sec:system_setup}. Then, we explain the proposed method and CNN structure and configuration in Section~\ref{sec:method} and the transfer learning procedure in Section~\ref{sec:training}. We provide experiments and results in Section~\ref{sec:results}, followed by relevant conclusions and future work in Section~\ref{sec:conclusion}.

\section{RELATED WORK}
\label{sec:related_work}

With the recent deep learning revolution in computer vision, especially for classification tasks, like ImageNet, it has been proven that it is possible to learn to identify objects in difficult environments and conditions~\cite{krizhevsky2012imagenet}.

In order to train a deep learning network, large amounts of training data are needed with precisely marked ground truth data. Collecting such training datasets can be a time-consuming task. However, transfer learning approach is useful when a fully trained system exists for one type of the problem and can be adapted for different datasets by adjusting some of the parameters of the network while keeping other parameters fixed~\cite{krizhevsky2012imagenet}. This has been proven to work for mid-level image representations in object classification, using the pre-trained network on natural images to adapt for medical image recognition and even emotion recognition~\cite{oquab2014learning}~\cite{greenspan2016guest}~\cite{ng2015deep}. Another interesting application of transfer learning is to use a fully trained network on night-time satellite imagery of poverty areas and adapt it to recognise poverty areas from daytime satellite imagery~\cite{xie2015transfer}. Furthermore, detailed analyses of the transfer learning approaches were made with surveys of the techniques used and various CNN structures~\cite{shin2016deep}~\cite{weiss2016survey}.

Moreover, CNN based work in the field of human pose estimation in 2D \cite{openpose2D2017}, known as {\it OpenPose}, allowed further improvements on 3D human pose estimation with the help of a depth sensor \cite{openpose3D2018}. The accuracy for a human keypoint in 3D is around $11cm$, mainly due to the inaccuracy of the depth sensor which grows with distance from the sensor.

On the other hand, many purely geometrical techniques have been employed to determine the position and orientation of an object from a single image by using some prior knowledge about the target object~\cite{cheng2011robust}~\cite{zhu2014single}.
In general, with these methods, they try to find patterns and features such as edges and corners which match the expected model and accordingly estimate the position and orientation. Some other researchers exploited the existence of a 3D model such as a CAD model~\cite{tsai2015cad}~\cite{lim2014fpm}
to increase the accuracy of the estimation. Although the precision of their method is higher compared to our CNN-based method, they mainly suffer from a major drawback: they can only perform with solid and rigid objects which clearly does not apply to robot manipulators. Another problem is the necessity of having a 3D model available beforehand, which in our method is substituted with the training procedure. However, our method performs more robustly in case of deviation from the model in case of physical damages or attached end-effectors, and it can also use the image colour information which is normally missing in a 3D model.

% Section System Setup
\section{SYSTEM SETUP AND DATASET COLLECTION}
\label{sec:system_setup}

Deep learning networks are capable of robustly recognising objects in complex backgrounds, but in order to achieve good performance, a large amount of precisely marked and diverse training data is needed. Considering the setup of three heterogeneous robotic manipulators, a system had to be set up to generate training data with accurate ground truth data marked automatically, given that manual ground truth generation for such datasets would take up a significant amount of time and effort.

Our setup consists of the following three robot types:
\begin{itemize}
  \item Universal Robots: UR3, UR5, UR10, 6 DoF, Figure~\ref{fig:ur}
  \item KUKA LBR iiwa - 7 R800, 7 DoF, Figure~\ref{fig:kuka}
  \item Franka Emika - Panda, 7 DoF, Figure~\ref{fig:franka_panda}
\end{itemize}

\begin{figure}[ht]
\vspace{-0.2cm}
    \centering
    \includegraphics[width=0.90\linewidth]{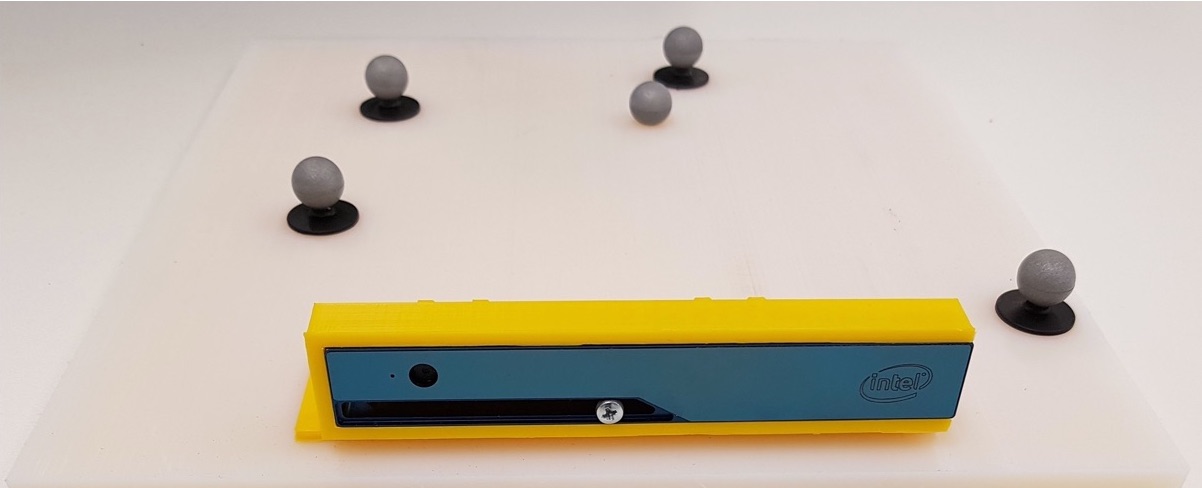}
    \caption{Setup with the Optitrack Motion Capture System}
    \label{fig:mocap_cam_setup}
\vspace{-0.2cm}
\end{figure}

This two-stage transfer learning work, as the basis, uses already trained multi-objective CNN \cite{miseikis2018multi}, which was trained on datasets containing all three robot models from UR. Datasets containing KUKA LBR iiwa were previously collected for our one-stage transfer learning project \cite{miseikis2018transfer}. All these datasets were collected using Kinect V2 sensor with necessary Eye-to-Hand calibration \cite{tsai1989new} every time position of the camera changed relative to the robot-base in order to achieve a high variety of backgrounds.

\begin{figure*}[ht]
\vspace{0.1cm}
\centering
    \includegraphics[width=0.80\linewidth]{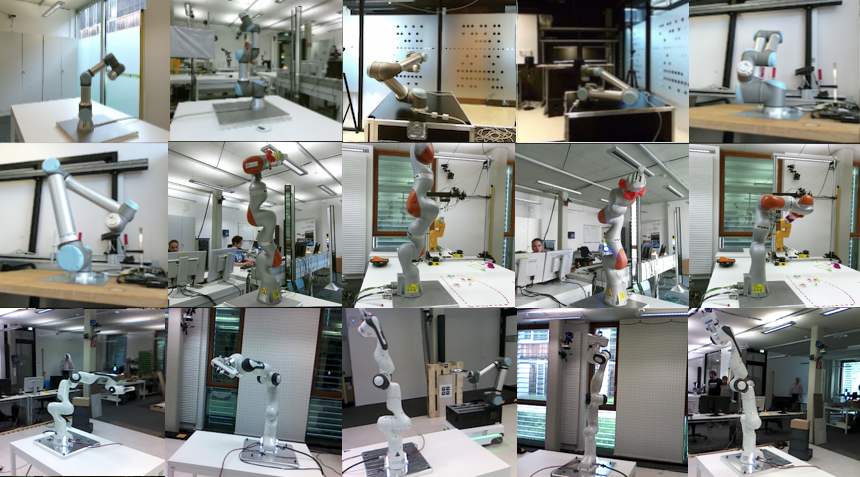}
    \caption{Samples from the collected robot datasets. Robots used are Universal Robots (silver-blue), Kuka LBR iiwa (silver-orange) and Franka Emika - Panda (white-black). A variety of robot configurations, camera movements and angles as well as lighting conditions were used. In some cases, even other robots in the background were present.}
    \label{fig:dataset_example_images}
\vspace{-0.7cm}
\end{figure*}

\begin{table}[h]
% \vspace{0.2cm}
\caption{Dataset summary describing a number of samples collected for each type of the robot. }
\label{table:dataset_summary_new}
\centering
\begin{tabular}{ |M{2cm}||M{2.4cm}|M{2.4cm}|}
 \hline
 Robot Type & Number of Datasets & Total Number of Samples \\
 \hline
 Universal Robots & $9$ & $4350$ \\
 Kuka LBR iiwa & $14$ & $1837$ \\
 Franka Panda & 5 & $2513$ \\
 \hline
\end{tabular}
\vspace{-0.2cm}
\end{table}

Table~\ref{table:dataset_summary_new} summarizes all datasets collected for each robot with their number of recordings where they differ by camera placement relative to the robot, illumination, background.

New datasets containing Franka Emika Panda robot were recorded with free-moving Intel RealSense R200 RGB-D camera instead of Kinect V2.
Since Eye-to-Hand calibration is only valid for the fixed camera setups, we could not use this method for the camera to robot coordinates-transformation measurements. 
Instead of performing Eye-to-Hand calibration, we have placed {\it Optitrack} (Motion Capture System)~\cite{point2011optitrack} markers over the moving camera and around the base of the robot in order to bring both systems into one coordinate frame of Optitrack (Figure~\ref{fig:mocap_cam_setup}). Since Optitrack's marker (\textit{Rigid-Body} or \textit{Rig}) origin is not exactly aligned with the camera's optical frame origin, extrinsic calibration was performed as described in \cite{chiodini2018camera} by observing and detecting one additional rig, that was fixed in the Optitrack frame, with our RGB camera from multiple positions. Example frames taken from the whole dataset can be seen in Figure~\ref{fig:dataset_example_images}.

Once all the transformations are connected in one coordinate system, a precise robot mask that is separating a robot body from the background when overlaying a colour image, used as a ground truth for teaching the CNN, can be calculated. It is done automatically by using ROS with {\it MoveIt} package \cite{sucan2013moveit}. The robot model, taken from the Robot Description Format (URDF) files and mesh files provided by the robot manufacturers, is updated with the live information from robot's joints encoder readings to create robot shape in real-time~\cite{meeussen2012urdf}. This shape is transformed to depth camera's coordinate frame and mask image is constructed.

\begin{figure*}[ht]
\vspace{0.1cm}
    \centering
    \includegraphics[width=0.80\textwidth]{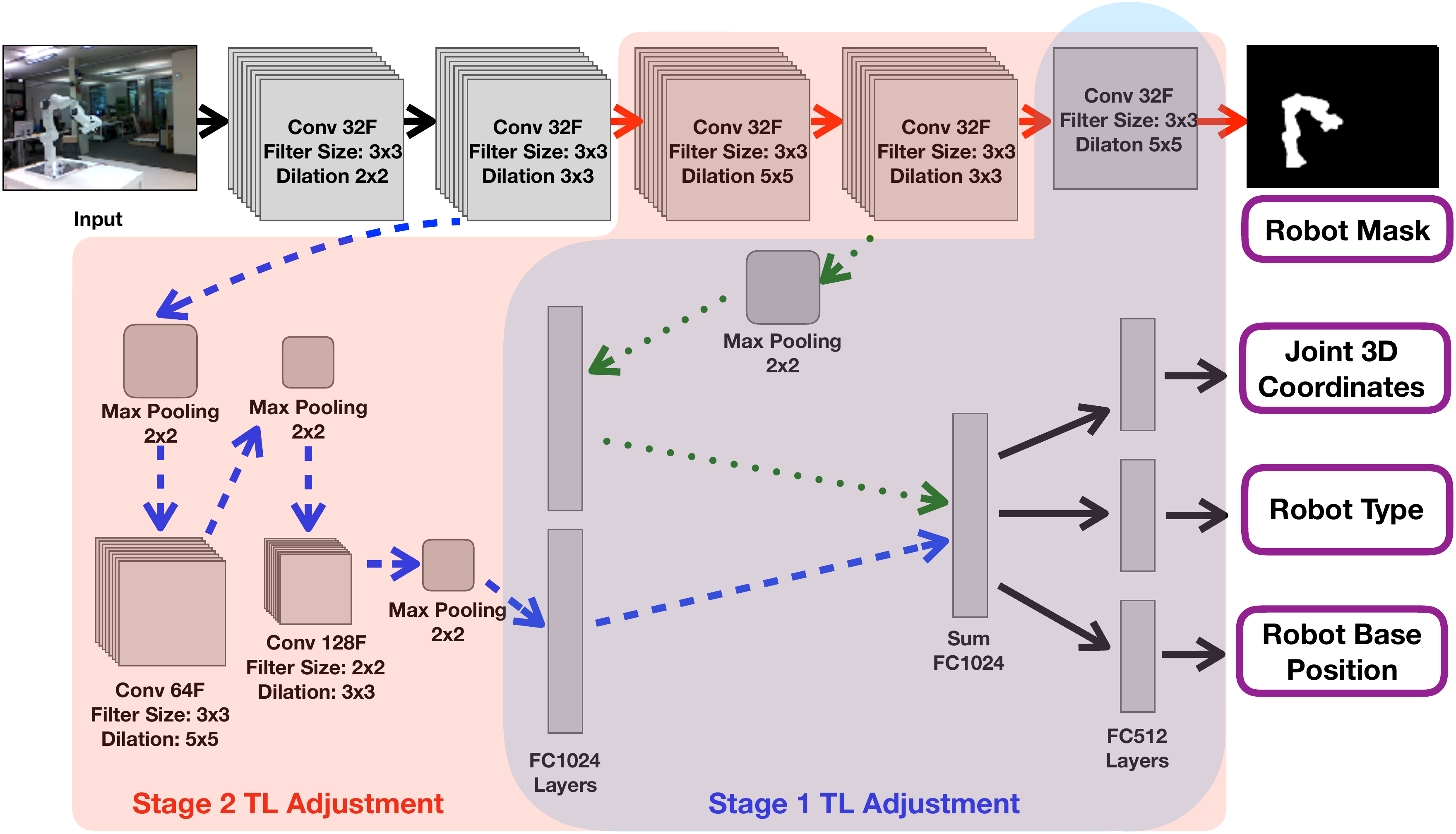}
    \caption{The multi-objective CNN with a two-stage transfer learning. The CNN is optimising simultaneously for four objectives at the same time: robot mask, 3D coordinates of robot joints, 3D coordinates of the robot base and the robot type. The network is taught in two stages using the transfer learning approach. In stage 1, the parameters for all the layers, besides the final ones marked in blue are \textit{frozen} and the system is trained until there is no more improvement. Afterwards, in stage 2, the parameters marked in red, as well as all the stage 1 layers, are adjusted during the training. This approach allows faster training compared to the full training, while still reaching good accuracy.}
    \label{fig:tf_cnn}
\vspace{-0.5cm}
\end{figure*}

In order to ensure the robustness of the system, robot movements were programmed so that each robot joint is moved through
the full range of motion in combination with other joints taking into account self-collision and table collision avoidance. Also, a trigger signal is used to save the data after each robot movement. With each trigger, we save camera colour images, robot joints and Cartesian coordinates, as well as ground-truth robot mask images. Moreover, to ensure a perfect overlap between colour and depth images, internal extrinsic camera calibration was used. All the input images are also rectified and have the resolution of $480 \times 360$ pixels. Testing and validation sets were divided by the ratios of $80\%$ and $20\%$ respectively based on random sampling.

% NEW Methods
\section{CNN STRUCTURE AND CONFIGURATION}
\label{sec:method}

The structure of a multi-objective CNN is identical to previous work, where it was trained on UR robots~\cite{miseikis2018multi}. The network trains for multiple outputs simultaneously by taking a single 2D colour image as an input. The network in this paper is trained on four objectives: Robot mask, Robot type, 3D Robot base position and 3D Position of the robot joints.

% The network in this paper is trained on four objectives:
% \begin{itemize}
%   \item Robot mask in the image
%   \item Robot type
%   \item 3D Robot base position
%   \item 3D Position of the robot joints
% \end{itemize}

The network has multiple branches, with some of the convolutional layers shared and then branched off to optimise specifically for each of the objectives. The structure of the multi-objective CNN can be seen in Figure~\ref{fig:tf_cnn}. 

\subsection{Loss Functions}

The loss function is needed to define the quality of training and drive the CNN towards achieving better results. Given four different outputs of the network, four loss functions need to be defined and eventually combined in a single one for the training process. Firstly, we describe each one of them and then explain how they are connected together.

Normally, the robot body takes up a rather small area in the whole image. For UR datasets, the area of the robot body is between $6-17\%$, for Kuka datasets, it is between $8-18\%$ and for Franka Panda, it is between $5-22\%$. Given a standard approach for the pixel classification loss function, an accuracy of over $78\%$ could be reached by classifying the whole image as background. This is conceptually wrong, so the loss function is adapted by calculating the foreground weight $w_{fg}$ as defined in Equation~\ref{eq:fg_weight}. It is based on the inverse probability of the foreground and background classes, where $Z \in \{fg, bg\}$.

\begin{equation}
    w_{fg} = \frac{1}{\prob(Z=fg)}
\label{eq:fg_weight}
\end{equation}

The background weight $w_{bg}$ is calculated in Equation~\ref{eq:bg_weight}.

\begin{equation}
    w_{bg} = \frac{1}{\prob(Z=bg)}
\label{eq:bg_weight}
\end{equation}

Definition of the loss function for the robot mask is done in two steps. First, a per-pixel loss $l^n$ is calculated in Equation~\ref{eq:loss_classification_pixel}, where $i_{est}$ is $\prob(Y = fg)$, $(1-i_{est})$ is $\prob(Y = bg)$ and $i_{gt}$ is the ground truth value from the mask image.

\begin{equation} \label{eq:loss_classification_pixel}
    \begin{split}
        l^n (I_{est}^n, I_{gt}^n) = 
        & -w_{fg} i_{gt} \log{(i_{est})} \\
        & - w_{bg}(1-i_{gt})\log{(1-i_{est})}
    \end{split}
\end{equation}

Then, a normalised loss calculation is done for the whole image $\mathcal{L}_{mask}$ in Equation~\ref{eq:loss_classification_full}. In order to keep the same learning parameters independent of the input image size, a normalisation factor $\mathcal{N}$ is used, which is a number of pixels in the image.

\begin{equation}  \label{eq:loss_classification_full}
    \mathcal{L}_{mask} (I_{est}, I_{gt}) = \frac{1}{\mathcal{N}} \sum\limits_{n} l^n (i_{est},  i_{gt})
\end{equation}

Estimation of the 3D coordinates of the robot base and robot joints are defined as a regression problem instead of classification. Loss function uses the Euclidean distance between the ground truth and estimated values by the CNN. For the robot joints estimation, the loss function $\mathcal{L}_{Jcoords}$ is described in Equation~\ref{eq:loss_joints_coords}, where $N_j$ is the number of joints, $J_{i}$ is the ground truth position of each joint and $E_{i}$ is the estimated values by the neural network.

\begin{equation} \label{eq:loss_joints_coords}
    \mathcal{L}_{Jcoords} = \frac{1}{N_j} \sum\limits_{i=1}^{N_j} \norm{J_{i}-E_{i}}_2
\end{equation}

\begin{figure*}[ht]
\vspace{0.1cm}
    \hfill
    \centering
    \begin{subfigure}[t]{0.32\textwidth}
        \centering
        \includegraphics[width=\linewidth]{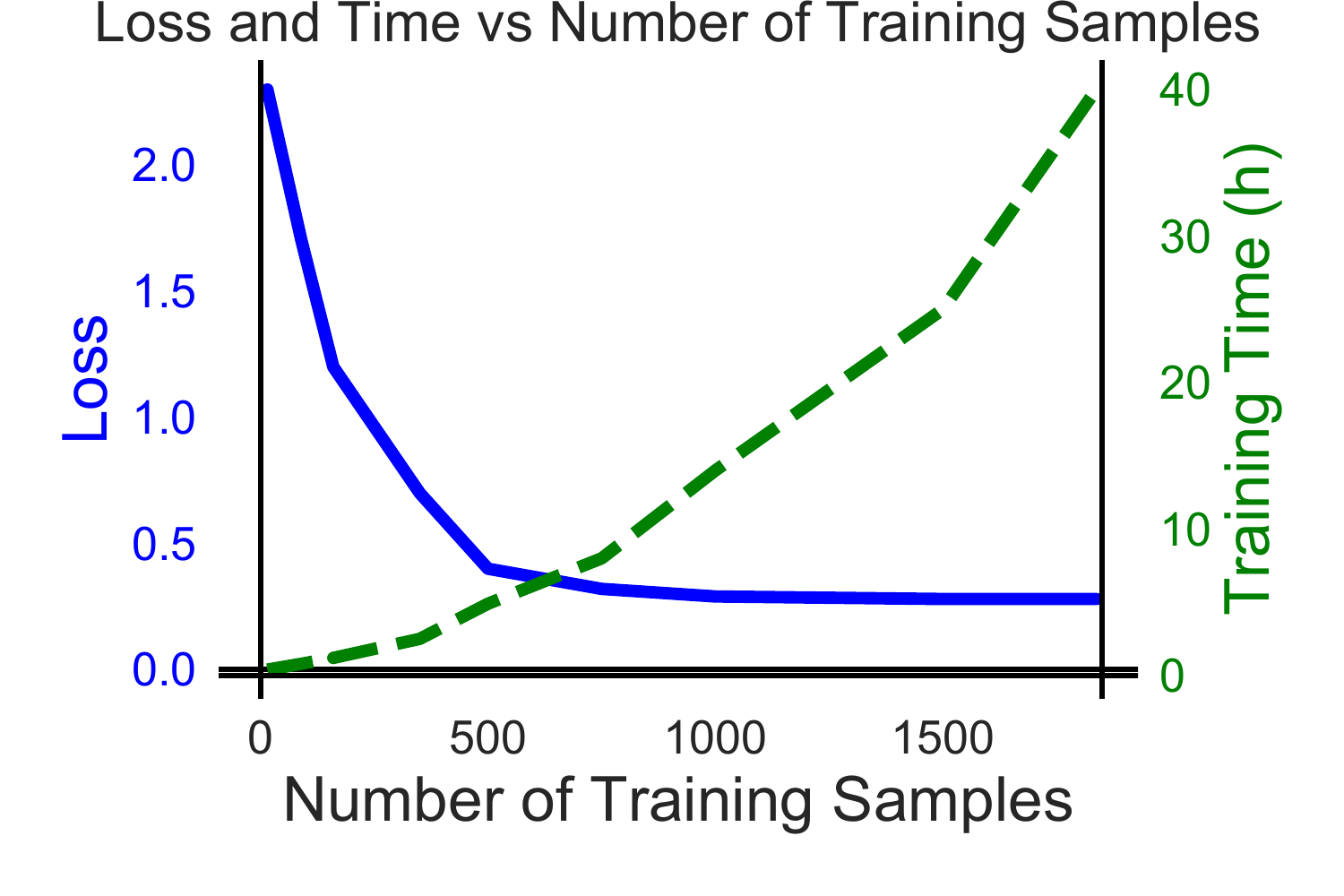}
        \caption{Loss function and training time against the number of training samples used. This was acquired by running a number of experiments using input datasets of different size. Using more than 500 training samples per robot type does not give a significant accuracy benefit while increasing the training time.}
        \label{fig:results_loss} 
    \end{subfigure}
    \hfill
    \begin{subfigure}[t]{0.32\textwidth}
        \centering
        \includegraphics[width=\linewidth]{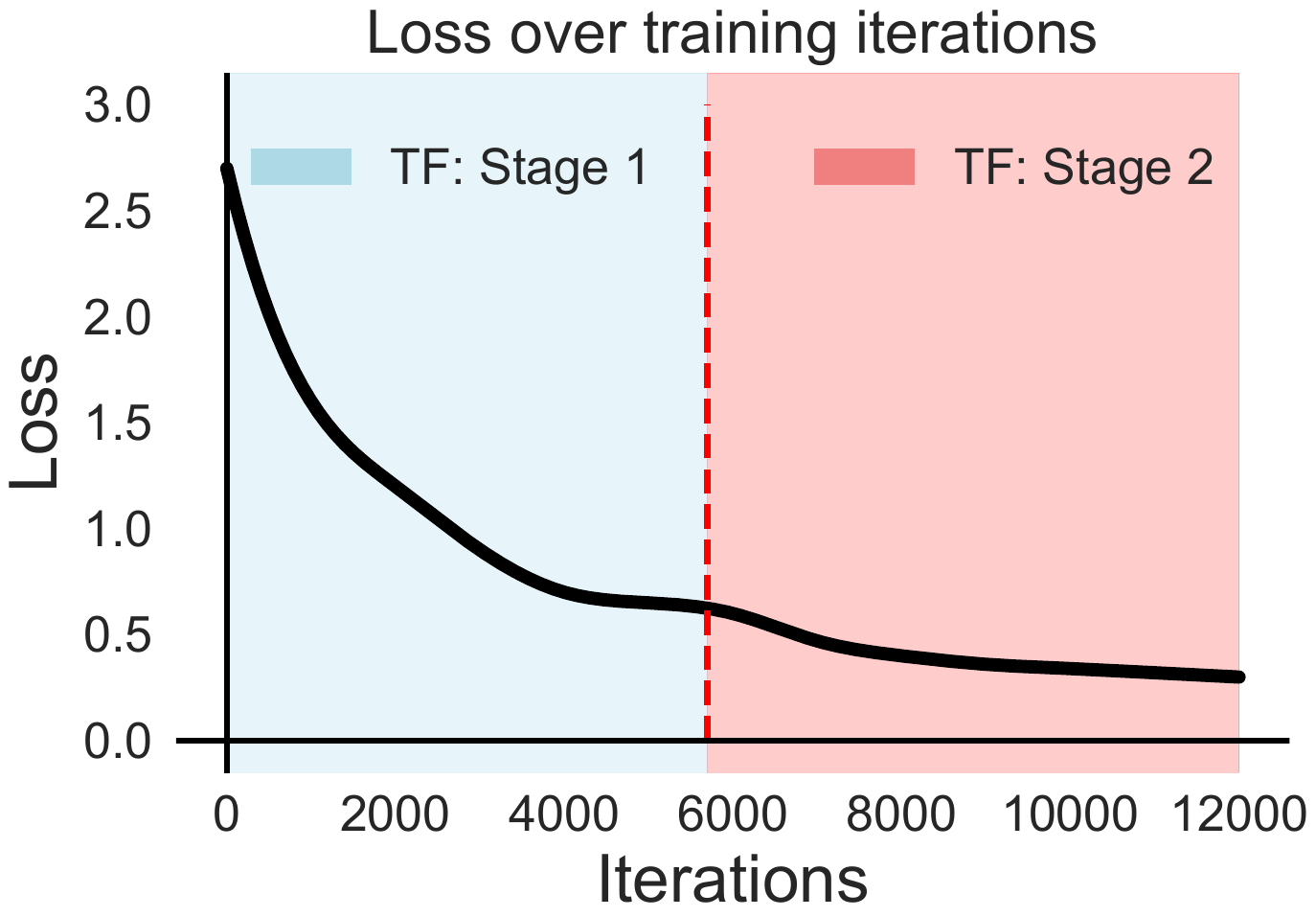}
        \caption{Loss function against the number of training iterations. Stage 1 and stage 2 of our two-stage transfer learning approach are marked by background colours on the graph. It can be seen that when stage 1 training saturates, unlocking parameters of more CNN layers allow the network to further improve results in stage 2 training.}
        \label{fig:results_stages}
    \end{subfigure}
    \hfill
    \begin{subfigure}[t]{0.32\textwidth}
        %\centering
        \includegraphics[width=\linewidth]{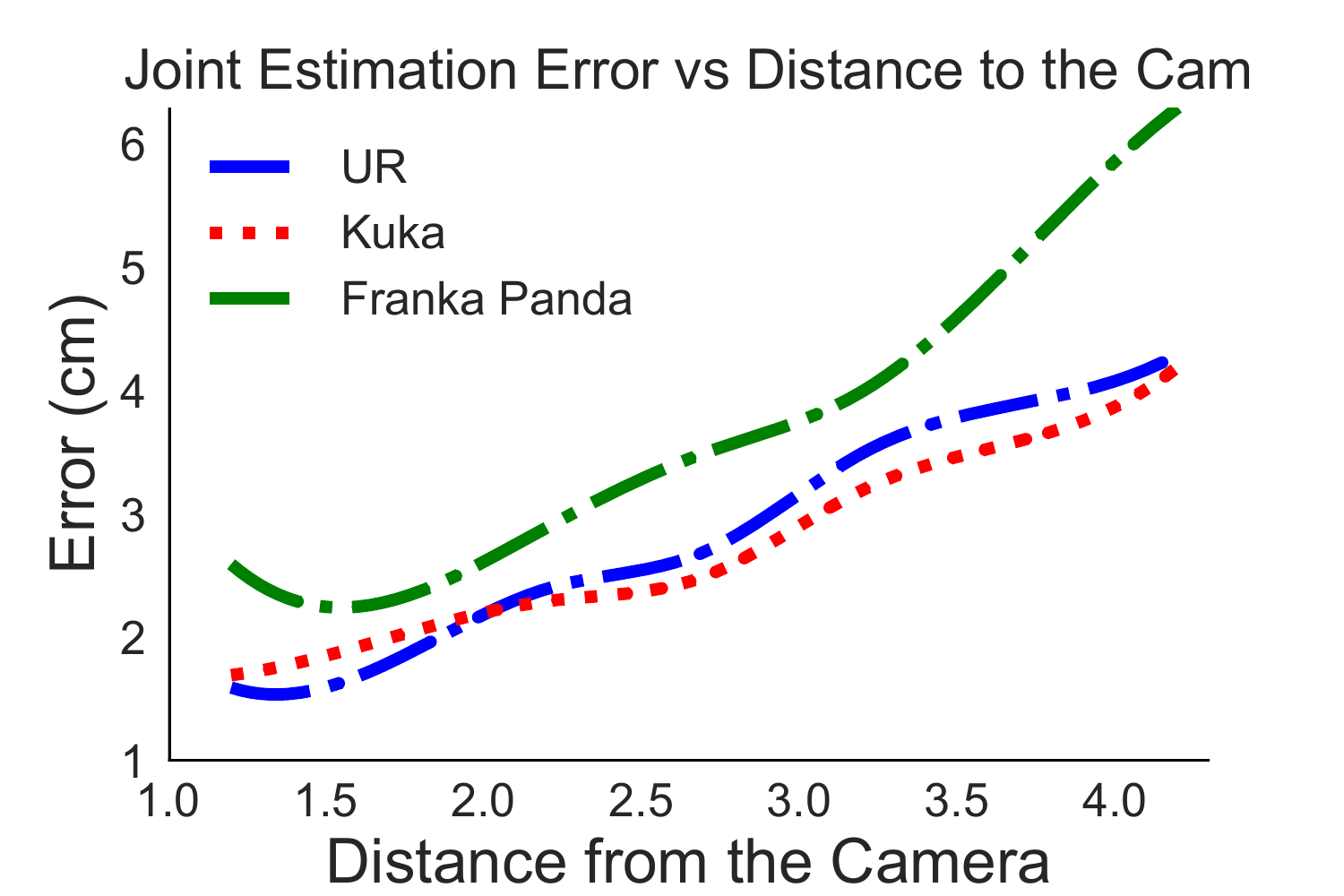}
        \caption{Errors for 3D position estimation of robot joints depending on the camera distance from the robot. Results are grouped by the robot type. Error and distance have close to linear correlation, but Franka Panda has higher error compared to the other robots. This is due to the robot body, which gives low contrast compared to the background.}
        \label{fig:results_distance}
    \end{subfigure}
    
    \caption{Evaluation of the two-stage transfer learning method using the test dataset in various categories.}
\vspace{-0.5cm}
\end{figure*}

The loss function for the coordinates of the robot base $\mathcal{L}_{Bcoords}$ is calculated in Equation~\ref{eq:loss_base_coords}. $B_{xyz}$ is the ground truth position of the robot base in 3D, and $E_{xyz}$ is the estimated 3D position of the robot base. These positions are relative to the coordinate frame of the camera.

\begin{equation} \label{eq:loss_base_coords}
    \mathcal{L}_{Bcoords} = \norm{B_{xyz}-E_{xyz}}_2
\end{equation}

A multi-class categorical cross-entropy approach is used to identify the robot type $\mathcal{L}_{type}$. $\mathcal{L}_{type}$ is calculated in Equation~\ref{eq:loss_robot_type}, where $p$ is the ground truth labels, $q$ are the predicted labels and $c\in R$, where $R$ contains all the available types of robots in the dataset.

\begin{equation} \label{eq:loss_robot_type}
    \mathcal{L}_{type} = -\sum\limits_{c} p(c) \log{q(c)}
\end{equation}

Eventually, all four previously defined loss functions are combined into a single loss function to be used in the training of the multi-objective CNN. The final loss function $\mathcal{L}_{final}$ is calculated as a weighted sum of all the loss functions, as shown in Equation~\ref{eq:final_loss}. The larger the weight $W$, the higher the impact on the corresponding value.

\begin{equation} \label{eq:final_loss}
    \begin{split}
        \mathcal{L}_{final} = 
        & W_{mask}\mathcal{L}_{mask} + W_{Jcoords}\mathcal{L}_{Jcoords} \\
        &+ W_{Bcoords}\mathcal{L}_{Bcoords} + W_{type}\mathcal{L}_{type}
    \end{split}
\end{equation}

\section{TRAINING USING TRANSFER LEARNING}
\label{sec:training}

A common approach to training such a system would be to train the whole system using full datasets of all the robots. However, this would take a significant amount of computation and time. Overall, the goal of this work is to analyse the possibility of having a pre-trained system and expand it to include more robot types while having a limited amount of training data and time.

Transfer learning allows us to use a fully trained system for one robot type and then adjust it to include the newly provided training data. This is done by \textit{freezing} the parameters in some of the layers while adjusting the remaining ones. Given this partial adjustment, the training time and amount of training data required can be significantly reduced.

The system was fully trained using UR robots with Kuka and Franka Panda robots added using the transfer learning approach. One crucial difference is that UR robots have 6 joints, while Kuka and Franka Panda are 7 joint robots. In general, it has been found that first convolutional layers tend to learn general visual features, while further layers figure out specific visual cues of the objects. Both, UR and Kuka robots have bright coloured joint covers, while the rest of the robot is silver, however, Franka Panda is mainly black and white, as seen in Figure~\ref{fig:robots_used}.

Due to these differences, a two-stage transfer learning approach was taken up, as shown in Figure~\ref{fig:tf_cnn}. In the first step, just the final layers of the multi-objective CNN are trained. This allows the network to adjust the dense layers to select the best-learned features for the robot recognition using currently learned visual cues. Only a small part of the CNN is adjusted, so the learning process is fast, and it re-learns robot classification and position estimation using current convolutional layer parameters.

However, after some point the learning process saturates and no more improvements are observed, defined by the reduction of loss. At this stage, the second part of the CNN is unlocked, allowing to modify parameters for the additional convolutional layers. This results in modifications of the visual cues that are learned as well as adjusting the final dense layers. The training speed is slower compared to the first stage of learning, but the loss is reduced even further.

In order to add the new robot types using the transfer learning approach, the new training dataset has to include both, the robot that the network was originally trained on, as well as the new robot(s) that should be recognised.

Weights for the loss function are adjusted to give more impact to identifying mask and robot type compared to our previous work. Selected weight values, based on trial and error from a number of experiments, were as follows:  $W_{mask}$: $1.2$, $W_{Jcoords}$: $1.2$, $W_{Bcoords}$: $1.2$ and $W_{type}$: $0.6$.
% \begin{itemize}
%     \item $W_{mask}$: $1.2$
%     \item $W_{Jcoords}$: $1.2$
%     \item $W_{Bcoords}$: $1.2$
%     \item $W_{type}$: $0.6$
% \end{itemize}

The number of training samples varied by the experiment and the input size of the images was scaled down and cropped to $256\times212$ pixels for all the datasets. The pixel intensity values of the input images were normalised to the range between 0 and 1. The learning rate was set to $0.001$ at the start of the training and then gradually decreased towards $0.000001$ as the training progressed.

% Section Experiments and Results
\section{EXPERIMENTS AND RESULTS}
\label{sec:results}

The main goal of the experiments was to evaluate the capability of including new robot types by using the two-stage transfer learning method while using a multi-objective CNN fully trained on UR robots as a starting point.

\begin{figure}[ht]
\vspace{-0.2cm}
    \centering
    \includegraphics[width=0.99\linewidth]{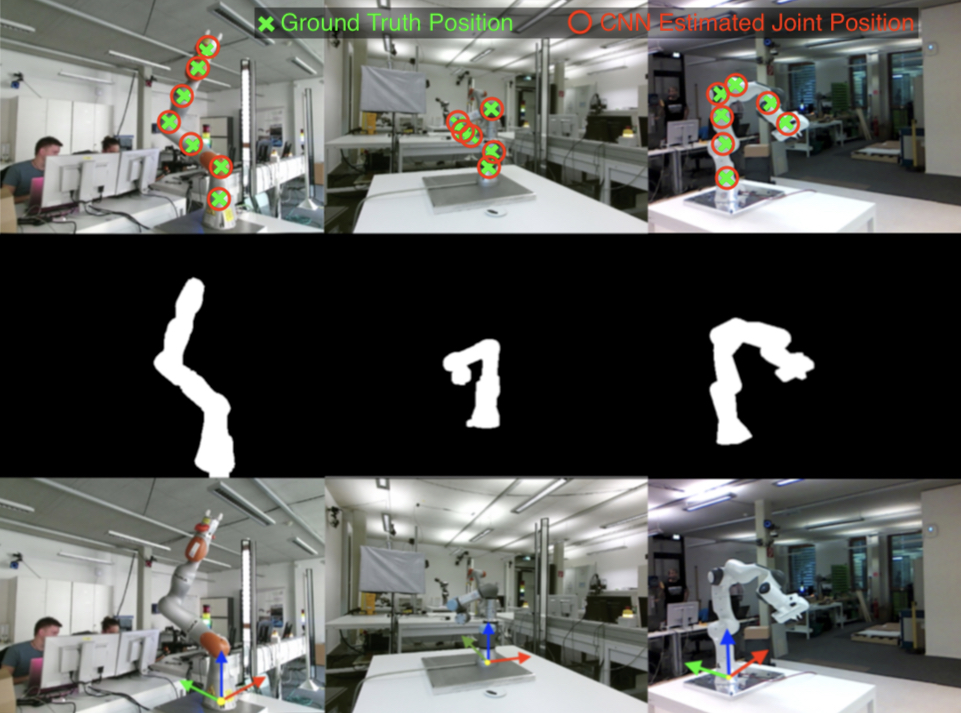}
    \caption{Visualisation of the output from the presented multi-objective CNN trained using a two-stage transfer learning approach. Each column represents each robot type in the following order: Kuka, UR and Franka Panda. The first row shows the estimated 3D joint position (red circles) against the ground truth position (green crosses). The second row shows the estimated mask of the robot and the third row shows the estimated robot base position.}
    \label{fig:marked_results}
\vspace{-0.4cm}
\end{figure}

Each of the experiments was conducted by taking a different size transfer learning dataset using a randomised sample selection to maximise the diversity of the data. The maximum amount of data was limited by the Kuka robot dataset to ensure the same amount of samples in each test for each of the robot types.

The system was evaluated using a testing set by comparing the output against the ground truth data. The robot mask accuracy is defined by counting the number of pixels in the CNN output image that match the ground truth mask. For the robot joint and base coordinates, the Euclidean distance between the CNN estimated results and ground truth results was calculated. We compare the results between each of the robot type in a number of categories. Results are summarised in Table~\ref{table:results_summary} and visualisation of estimations plotted on top of the testing set samples can be seen in Figure~\ref{fig:marked_results}.

\begin{table}[h]
\vspace{-0.2cm}
\caption{Summary of the Transfer Learning results (using 500 samples of each robot type for training) on the test set.}
\label{table:results_summary}
\centering
%\begin{tabular}{ | c || c | c | c | c | }
\begin{tabular}{ |p{3.1cm}||p{1cm}|p{1cm}|p{1.6cm}|}
\hline
Measure & UR & Kuka & Franka Panda \\
 \hline
 Mask Accuracy, \% & $97.0\%$ & \boldmath$97.7\%$ & $94.3\%$ \\
 Robot Type Accuracy, \%  & $97.5\%$ & \boldmath$100\%$ & $96.1\%$ \\
 Joint Pos Error (Median) & \boldmath$3.12 cm$ & $3.30 cm$ & $3.64 cm$ \\
 Base Pos Error (Median) & $2.42 cm$ & \boldmath$2.36 cm$ & $3.01 cm$ \\
 \hline
\end{tabular}
\vspace{-0.3cm}
\end{table}

All of the robots had joint 3D position estimation error under $3.64 cm$, with the base position estimation error under $3 cm$. The mask accuracy estimation exceeded $97 \%$ for UR and Kuka robots, while for Franka Panda it was a bit lower at $94.3 \%$. Robot type was recognised correctly in all of the cases for Kuka robot, while UR and Franka Panda recognition were $97.5 \%$ and $96.1 \%$ respectively. Overall, it was noticed that given the distinct features, the CNN performed the best on Kuka robot, while low contrast Franka Panda robot had the worst results, but not far behind.

Considering overall performance of the two-stage transfer learning, as shown in Figure~\ref{fig:results_loss}, for the multi-objective CNN, it can be seen that the loss function stops improving when having datasets of size between $500$ and $750$ training samples for each of the robot type, which corresponds to $7$ to $10$ hours of training time. Increasing the number of training samples beyond $750$ does not improve the learning process, but significantly increases the training time.

Compared to the previously presented work in~\cite{miseikis2018multi}, the detection accuracy of the current two-stage transfer learning approach achieved similar accuracy in a joint position error of $2.46 cm$ vs $3.12 cm$ and slightly worse accuracy for the robot mask estimation: $97\%$ vs $98\%$ in the previous work. The full training time of the multi-objective CNN for UR robots took $60$ hours vs $10$ hours in the current work.

The performance of each training stage of transfer learning is shown in Figure~\ref{fig:results_stages}. Stage 1, with parameters in final CNN layers being adjusted, saturates after $6000$ iterations. Afterwards, further layers are \textit{unlocked} switching to Stage 2 and the loss function reduces even further settling down between $10000-12000$ iterations.

Furthermore, we analyse the impact of the joint and base position estimation depending on the distance between the camera and the robot, visualised in Figure~\ref{fig:results_distance}. There is close to a linear relationship between the distance between the robot and the accuracy of the 3D position estimation of the robot joints. Interestingly, at a very close distance of 1.2 meters, Franka Panda robot shows worse performance compared to 1.5 meter distance. %However, this is likely to be caused by having parts of the robot not fitting into the field of view of the camera when freely moving the camera around in the training dataset.

The detection time or forward-propagation of the multi-objective CNN was measured to be $19-23 ms$ per frame on a nVidia GTX 1080 Ti graphics card.

\section{CONCLUSIONS AND FUTURE WORK}
\label{sec:conclusion}

In this paper, we have presented a two-stage transfer learning approach, which allows to re-train a previously trained multi-objective CNN to include numerous new robot types using a limited amount of training data. This approach reduces the time spent on collecting datasets with ground truth data, as well as the training time of the network itself. Furthermore, a concept of a multi-objective CNN capable of identifying heterogeneous robots, classifying their types and estimating 3D positions of their joints and base was proven. A simple 2D colour image was used as an input and % from two different vision sensors, Intel RealSense and Kinect V2 with depth information omitted in the actual application. 
Kuka and Franka Panda robots were mounted with two-finger grippers on the end-effector, which were not taken into account by the CNN. The network successfully estimated the position of the robot as the was no end-effector present. If the TCP of the end-effector would be required, it could be calculated by adding the necessary CAD model or offset information to the estimated position of the end-effector.

With the detection time of under $23 ms$, the system has proven to be capable of working in real-time. At the current stage, a powerful GPU is needed to run it, however, a goal of optimising it for smaller mobile systems could be pursued. In this case, it could be implemented in small wearable cameras to be used both, for mobile robots or for human operators working in a robotised environments and used as a safety system, which can detect possible collisions without having any direct communication between the devices. The outcome could be a valuable measure for various safety applications in {\it Human-Robot Interaction} scenarios, where we need to know the position of the human and robot and their individual joints respective to each other.

The achieved robot joint position estimation is not accurate enough for visual servoing operations, but future work can focus on accuracy improvements. We believe that by using higher resolution images, multi-sensor detection and tracking in time series, the accuracy of our system could be improved. Furthermore, an analysis of the impact on the detection accuracy depending on the weight selection of loss functions and layer selection for the transfer learning will be done. With the current results, a high-level control is still possible for human-robot and robot-robot interaction.

Additionally, with the given robot mask detection in a 2D image, some robot self-inspection could be done to detect any damage, especially for autonomous robots operating in remote or disaster areas, where people do not have access to, for example for planetary exploration rovers.% By further training this system on commonly observed damaged textures, it can be capable of marking and assessing the severity of the damage.

% \subsection{FUTURE WORK}

% For most of the robotic applications that we need to calculate the relative position of a robot to a human, the distance accuracy of a few centimeters may be sufficient. If a human is approaching a robot from a few meters distance, then a few centimeters of error is not that critical. However, with advances in collaborative robots, some applications have arisen that need a very close cooperative work between robot and human. For such applications, we can also focus on the improvement of our method by:
% \begin{itemize}
%   \item Working on higher resolution images.
%   \item Using multiple cameras, or stereo vision to acquire more reliable depth information.
%   \item Perform a Eye-to-Hand calibration in advance, between the camera and the robot, to measure the position of the Optitrack markers with a higher accuracy.
%   \item Combine this method of robot localization with human localization so that we can mount the cameras around the workspace without the need of mounting it directly on humans.
% \end{itemize}
 % End of blue coloured text